\documentclass[conference]{IEEEtran}
\IEEEoverridecommandlockouts
\usepackage{cite}
\usepackage{amsmath,amssymb,amsfonts}
\usepackage{algorithmic}
\usepackage{graphicx}
\usepackage{textcomp}
\usepackage{xcolor}

\usepackage{mdframed}
\usepackage{hyperref}
\usepackage{fancyvrb}
\usepackage{array}
\usepackage{tabularx}
\usepackage{array}
\usepackage{booktabs} 
\usepackage{float}
\usepackage{util}

\def\BibTeX{{\rm B\kern-.05em{\sc i\kern-.025em b}\kern-.08em
    T\kern-.1667em\lower.7ex\hbox{E}\kern-.125emX}}
\begin{document}
{


\title{Multi-task Prompt Words Learning for Social Media Content Generation
}

\author{
    \IEEEauthorblockN{Haochen Xue$^{*}$ \thanks{* Equal Contribution.
    $\dagger$ Corresponding Authors Xiaobo Jin: Xiaobo.Jin@xjtlu.edu.cn. This work was partially supported by the “Qing Lan Project” in Jiangsu universities, Research Development Fund with No. RDF-22-01-020 and Suzhou Science and Technology Development Planning Programme (Grant No. ZXL2023176)}, Chong Zhang$^{*}$, Chenzhi Liu$^{*}$, Fangyu Wu, Xiaobo Jin$^{\dagger}$}
    \IEEEauthorblockA{
        School of Advanced Technology, Xi'an Jiaotong-Liverpool University, Suzhou, China \\
    }
    \IEEEauthorblockA{
        \textbf{Email:} \{Haochen.Xue20, Chong.Zhang19, Chengzhi.Liu21\}@student.xjtlu.edu.cn, \\
        Fangyu.Wu02@xjtlu.edu.cn, Xiaobo.Jin@xjtlu.edu.cn
    }
}

}

\maketitle

\begin{abstract}

The rapid development of the Internet has profoundly changed human life. Humans are increasingly expressing themselves and interacting with others on social media platforms. However, although artificial intelligence technology has been widely used in many aspects of life, its application in social media content creation is still blank. To solve this problem, we propose a new prompt word generation framework based on multi-modal information fusion, which combines multiple tasks including topic classification, sentiment analysis, scene recognition and keyword extraction to generate more comprehensive prompt words. Subsequently, we use a template containing a set of prompt words to guide ChatGPT to generate high-quality tweets. Furthermore, in the absence of effective and objective evaluation criteria in the field of content generation, we use the ChatGPT tool to evaluate the results generated by the algorithm, making large-scale evaluation of content generation algorithms possible. Evaluation results on extensive content generation demonstrate that our cue word generation framework generates higher quality content compared to manual methods and other cueing techniques, while topic classification, sentiment analysis, and scene recognition significantly enhance content clarity and its consistency with the image.

\end{abstract}


\section{Introduction}

Social media plays a vital role in today's world, not only as a key platform for human connection, information sharing and cultural interaction but also provides individuals with the opportunity to showcase themselves and benefit on both a personal and professional level \cite{kaplan2010users}. It also provides companies with effective tools to enhance their marketing strategies and increase product exposure. Therefore, creating high-quality tweets can significantly enhance an individual's social media presence and personal brand, and increase engagement and connection with followers, promoting the sharing and exchange of ideas and opinions. For enterprises, high-quality tweets can effectively attract more attention, strengthen brand image and trust, and effectively disseminate information and promotions to target audiences, thereby promoting business development and market expansion \cite{alves2016social}.

However, despite the rapid advancement of artificial intelligence technology in content creation, particularly with the use of GPT for text production \cite{davis2022investigating}, its application in social media remains underdeveloped. Besides the issue, for users, especially those new to social media, creating visually appealing photos and engaging text content can be challenging \cite{carr2015social}. Tasks like photo editing and making captions can be difficult even for skilled users. In general, editing pictures and content is a difficult and time-consuming job for ordinary users. Bridging these gaps, the application of the technology not only holds the potential to unlock new avenues for innovation and diversity in social media content but also brings convenience to users. 

The goal of our work is to achieve controllable automatic generation of social media content through existing multi-modal artificial intelligence technology. Specifically, we first generate a preliminary content description of the image based on multi-modal technology. Next, based on the generated content and original images, we perform our multi-task prompt words learning (MPWL): topic classification, sentiment analysis, scene recognition and content keyword extraction. These four types of prompt words are then fed into GPT through specific templates to optimize GPT's performance on social media generation tasks. With the help of MPWL and ChatGPT, content can be generated that is concise, smooth, consistent with pictures, and in line with human aesthetics. Finally, the GroundingDINO \cite{liu2023grounding} algorithm is applied to detect people in the image so that the cropped image is centered on the person. Our work will provide tool convenience, content diversity, and greater time efficiency for content creation in social media in the future.

Our main contributions are as follows:

\begin{itemize}

\item We have applied multimodal processing and large language model to the field of social media content generation. As far as we show, we are the first work to use these techniques for the automatic generation of social media content.

\item We propose a generation framework of prompt words based on multi-modal information fusion, which combines tasks such as topic classification, sentiment analysis, scene  recognition and keyword extraction to generate more comprehensive prompt words for social media to guide and control ChatGPT to generate high-quality of tweets.

\item In the current lack of evaluation tools and standards in the field of content generation, we use the ChatGPT tool to make it possible to evaluate experimental results on a large scale, which is more standardized and fair than manual evaluation.

\item In our experiments, the tweets generated by our just-in-time learning method were of higher quality than manually crafted tweets, while also outperforming tweets generated by other just-in-time learning methods.

\end{itemize}

\section{Related Work}

\subsection{Artificial Intelligence-Generated Content}
The research field of Artificial Intelligence-Generated Content (AIGC) has witnessed a surge in interest, particularly for its remarkable capacity to autonomously create diverse content forms, encompassing text, images, and videos \cite{yunjiu2022artificial}. Within the domain of text generation, researchers have delved into sophisticated methods utilizing deep learning techniques to craft coherent and varied texts, as evidenced by studies such as \cite{karras2017progressive} and \cite{huang2018multimodal}. A pivotal development in the AIGC landscape is the emergence of Large Language Models (LLMs), exemplified by GPT-3 \cite{brown2020language}, built on the transformative Transformer architecture. These models, trained on massive datasets, embody extensive general world knowledge and exhibit exceptional prowess in natural language processing tasks. Distinguished by their large-scale training datasets and a significant number of parameters, LLMs excel in capturing intricate linguistic patterns and nuances. Their effectiveness spans a spectrum of tasks, including natural language inference, question answering, and code generation  \cite{chowdhery2023palm, chung2022scaling, poesia2022synchromesh}. 

Recently, researchers have leveraged the few-shot capabilities of LLMs across diverse applications, such as translation, text summarization, and common sense reasoning \cite{brooks2023instructpix2pix, schick2023toolformer}. Brooks et al. \cite{brooks2023instructpix2pix} harnessed GPT-3 for generating instructions and editing captions datasets, contributing to the training of models for image editing. Schick et al. \cite{schick2023toolformer} utilized GPT-3's few-shot capability to curate a dataset of external API calls, subsequently fine-tuning another LLM. In a similar vein, Peng et al. \cite{peng2023instruction} utilized GPT-4 outputs to construct an instruction-following dataset, facilitating the subsequent fine-tuning of other LLMs in a supervised learning fashion.

Despite these advancements, the field of generative AI faces challenges, especially in the intricate domain of social media content generation, where tasks extend beyond text to include the creation of graphics. Additionally, controlling generative AI remains a formidable task, given that text prompts serve as the primary interface. While re-prompting in a chat-like environment offers a method for refining outputs \cite{openai2023gpt4}, the precise identification of issues and corresponding solutions remains a challenge. To address these challenges, this work proposes a novel Multi-task Prompt Word Learning approach. This approach aims to enhance control over the generation process of GPT and elevate the quality of generated social media content.

\subsection{Prompt Learning}
To adapt pre-trained Large Language Models (LLMs) for downstream tasks, prompt learning strategies \cite{gan2022vision, liu2023pre} traditionally employ task-specific textual tokens to elicit task-specific textual knowledge \cite{liu2023pre, radford2021learning}. For instance, in CLIP \cite{radford2021learning}, a manually crafted template like "a photo of a [CLASS]" is employed to construct textual embeddings for zero-shot prediction. However, these manually designed prompts exhibit limitations in offering comprehensive descriptions of downstream tasks as they lack consideration for specific knowledge pertinent to the task at hand.

To overcome this limitation, Context Optimization (CoOp) \cite{zhou2022learning} replaces manually crafted prompts with learnable soft prompts inferred from labeled few-shot samples. However, CoOp's drawback lies in generating unique and fixed learnable prompts for each task's images, neglecting the nuanced characteristics of individual images. In response, Conditional Context Optimization (CoCoOp) \cite{zhou2022conditional} is introduced, generating image-conditional context for each image and amalgamating it with textual-conditional context for prompt tuning. Specifically, a lightweight neural network is employed to generate a vector representing learnable text prompts. ProDA \cite{lu2022prompt}, in its pursuit of high-quality task-related tokens, incorporates prompt prior distribution learning. Additionally, ProGrad \cite{zhu2023prompt} selectively updates prompts aligned with the gradient of the "general knowledge" generated by the original prompts.

Graph-ToolFormer \cite{zhang2023graph} significantly enhances LLMs' capabilities in graph reasoning tasks. To enhance efficiency and alleviate computational load in LLMs, LLMLingua \cite{jiang2023llmlingua} integrates budget controllers, iterative token-level compression, and directive-based adjustments. Prompt Distillation \cite{li2023prompt} transforms discrete prompt words (traditional text prompts) into continuous prompt vectors, reducing computational load during input processing, as continuous vectors are more amenable to the model compared to traditional text. DenseCLIP \cite{rao2022denseclip} employs a context-aware prompt strategy for generating dense prediction tasks, while CLIP-Adapter \cite{gao2023clip} applies an adapter to fine-tune visual or text embeddings.


\section{Methodology}

\begin{figure*}[!htp]
    \centering
    \vspace{-10pt}
    \includegraphics[width=1.05\textwidth]{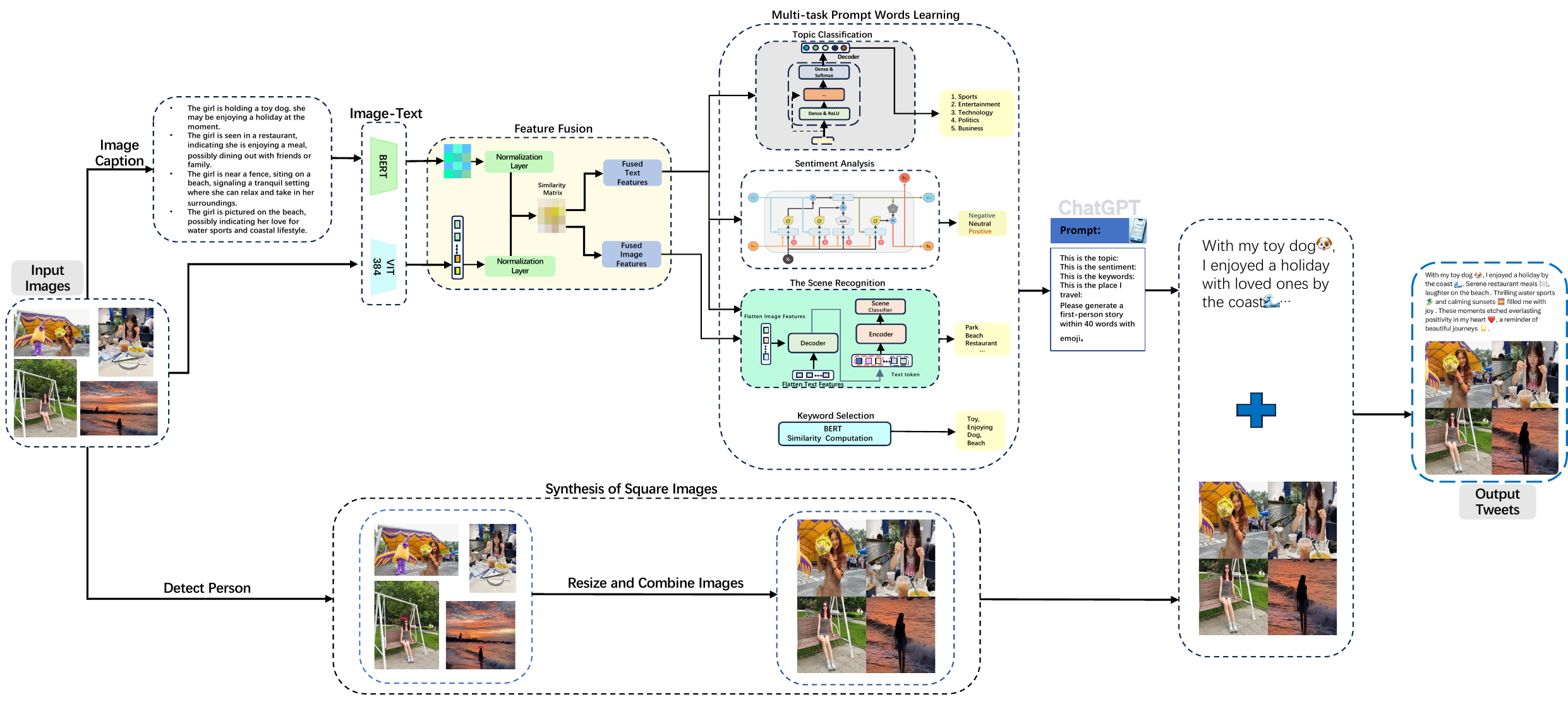}
    \caption{The pipeline of our algorithm is as follows: 1) Image features and text features are fused after BERT and VIT; 2) Based on the fused image features and text features, we use multi-task learning to generate prompt words; 3) Twitter text is generated based on ChatGPT and a template filled with prompt words; 4) The synthesized square image and Twitter text are combined into the final tweet containing images and text.}
\label{fig:example}
\end{figure*}

The framework of our algorithm is as follows: 1) Image features and text features are fused after BERT and VIT; 2) With the fused image features and text features, we use multi-task learning including topic classification, sentiment analysis, scene recognition and keyword extraction to generate prompt words; 3) Twitter text is generated based on ChatGPT and a template filled with prompt words; 4) The synthesized square image and Twitter text are combined into the final tweet containing images and text. Below we introduce each component of the algorithm in detail.

\subsection{Crude Content Generation}

It is first processed using a ViT ImageProcessor to convert these images into a vectorized semantic representation. These vectors are then fed into HuggingGPT \cite{shen2023hugginggpt}, an image captioning model that is trained to encode and decode visual information using ViT and GPT-2’s \cite{khan2022transformers} causal language model, respectively. Through the above process, a preliminary text description is generated for each image patch. It is important to note that we improve the accuracy of the model by fine-tuning the COCO dataset\cite{lin2014microsoft}. Through the inherent randomness of the model and the adopted Top-k decoding strategy, we improve the diversity of the generated text, producing several possible descriptions for each image.

In order to get a more accurate text description of the image, we first generate $4$ random text descriptions for the image, then use CLIP (Contrastive Language-Image Pre-training) \cite{Li_2022_CVPR} to score the matching of each text and image. Finally, the text description with the highest score is selected as the description of the image. CLIP here is a text-to-image pre-trained model designed to learn alignment between multi-modal entities such as images and text. Given raw inputs of text and images, CLIP will output a similarity representation between them, where $-1$ means completely dissimilar and $1$ means completely similar.

\subsection{Multi-task Prompt Words Learning}

In this section, we will generate image-aligned text, specifically using large language cue learning methods to generate textual descriptions that are consistent with the visualization's context. Prompt learning is a creative and detail-oriented task that involves strategic choices of words, phrases, symbols, and formatting, all designed to guide the model in producing high-quality and relevant text.

\begin{figure}[ht]
    \centering
    \includegraphics[width=0.48\textwidth, height = 3.3cm]{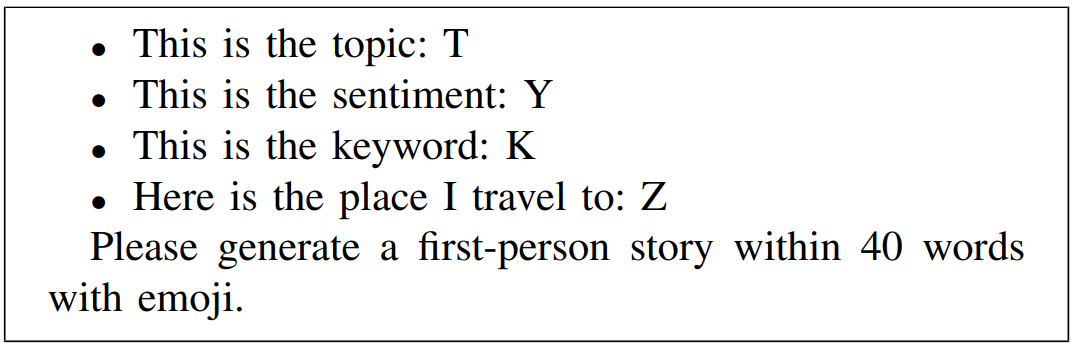}
    \caption{Prompt word template for multi-prompt word learning}
    \label{prompt1}
\end{figure}

To this end, we generate prompt words through the following multiple methods: 1) extract keywords $K = \{k_1, k_2, \ldots, k_m\}$ from a series of candidate sentences; 2) classify multi-modal data to determine the topic words $T$ of the text; 3) perform sentiment analysis on text data and evaluate their emotional polarity $Y$ ($Y$ may be positive, negative and neutral); 4) identify the scene $Z$ in which the story takes place through text and images. Subsequently, we will use the prompt template of Figure \ref{prompt1} to synthesize the final prompt text description with the above information $T$, $Y$, $K$ and $Z$.


\paragraph{Keyword  Extraction}  After obtaining the text description of the image, we first perform part-of-speech filtering on all words in the text: analyze the part-of-speech of each word with the help of CoreNLP \cite{manning2014stanford} and retain semantic words such as adjectives, names, and verbs as candidates $\mc{W}$. The final keywords will be selected from these words.

The next step is to measure the importance of keywords: mask the sentences in the paragraph and calculate the distance between the two sentences before and after the masking. Given a word sequence corresponding to a sentence $S = (w_1,w_2,\cdots,w_k,\cdots,w_n)$, we encode it into a 768-dimensional vector through BERT\cite{devlin2018bert} and then calculate the cos similarity between words degree sentence and its masked version to measure the importance of word $w_k$
\begin{equation}
h(w_k) = \frac{\tm{BERT}(S)^T \tm{BERT}(S/w_k)}{\|\tm{BERT}(S)\|_2 \|\tm{BERT}(S/w_k)\|_2},
\label{eq:keyword_importance}
\end{equation}
where $S/w_k$ represents the sentence after removing (masking) the word $w_k$. Here, $h(w_k)$ basically measures the similarity distance between two sentences. The smaller the distance $h(w_k)$, the more important the word $w_k$ is. Finally, we select the most important 3 words from the candidate words of the sentence as keywords.

\paragraph{Fusion of Multimodal Features}

In this section, we attempt to fuse the feature representations of multiple models to generate more accurate prompt representations that will support subsequent multiple learning tasks.

To achieve this fusion effectively, we begin by utilizing specific pre-trained models for distinct modalities. We use pre-trained ViT-384 \cite{han2022survey} as the backbone to extract image features and pre-trained BERT as the backbone to extract text features. 

Subsequently, we normalize all image features and word-level features to obtain matrices $I_{l \times m} = \bmtx{i_1 & i_2 & \cdots &i_m}$ and $W_{l \times n} = \bmtx{w_1 & w_2 & \cdots & w_n}$, where vectors $i_j$ and $w_j$ are unit vectors in $l$-dimensional space respectively.

In order to better integrate the semantic representation of text and images, we use the idea of Transformer to regenerate the representation of text and images
\eqna{
\mc{C} & = & \tm{softmax}\left( \frac{I^T W}{\sqrt{l}} \right) \\
\widehat{I} & = & W \mc{C}^T, \\
\widehat{W} & = & I \mc{C}.
}

Here, $\widehat{I}$ and $\widehat{W}$ will be used as inputs for multiple subsequent tasks including topic classification, sentiment analysis and scene recognition.

\paragraph{Topic Categorization} In order to ensure the diversity of GPT model generation, we will build a topic model to classify all samples into five categories: sports, entertainment, technology, politics, and business. We used a web crawler to crawl 1500 samples containing images and text from Twitter to build our model, where the training, validation, and test sets were randomly divided in 80:10:10.

\begin{figure}[ht]
    \centering
    \includegraphics[width=0.26\textwidth, height = 7cm]{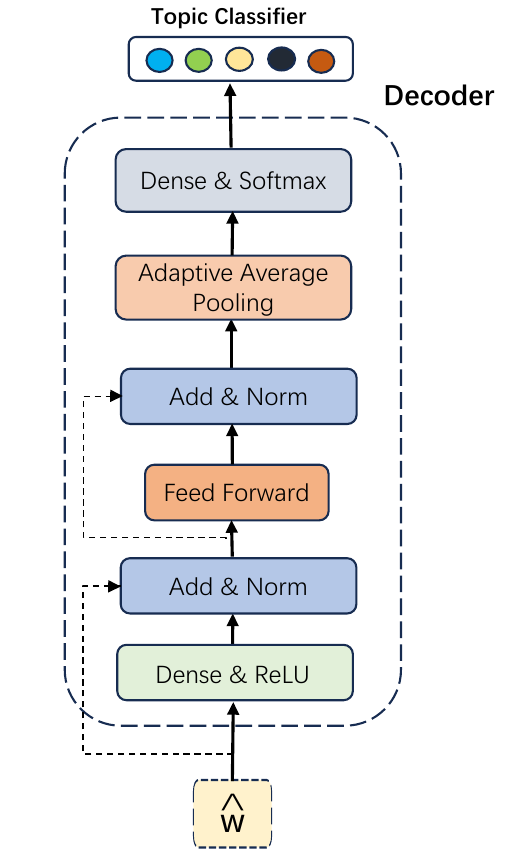}
    \caption{Network architecture for topic analysis task}
    \label{fig:topic}
\end{figure}

In order to process $\widehat{W}$, we design a decoder to complete the task of topic classification, as shown in Figure \ref{fig:topic}. The Dense/ReLU layer initially processes the fused features, providing basic nonlinearity. Next, the Add/Norm layer with its residual connections and normalization ensures the stability and depth of feature learning. Next comes the Feed Forward layer, which further enhances the model’s ability to capture complex patterns. Adaptive average pooling adjusts feature sizes for efficient processing. Finally, a combination of dense layers and softmax functions converts these processed features into classification probability distributions, which are crucial for accurate topic classification.


\paragraph{Sentiment Analysis} 

Below we predict the sentiment polarity of the semantics of image and text representations and use it in prompt templates.

For text sequence data, we use LSTM to predict the sentiment polarity of the text $\widehat{W}$ integrated into the image. LSTM contains LSTM cells that process the input data of each time step in a recurrent fashion.
LSTM cells include forget gates, input gates, output gates, cell states and hidden states.
The forget gate, input gate and output gate in Eqn. (\ref{fgate}), (\ref{igate}) and (\ref{ogate}) can operate the input data $z_t$ and the hidden state $h_{t-1}$ to obtain the corresponding gating coefficients, where $\sigma$ is the sigmoid activation function, and $\bm{W}$ and $\bm{b}$ are weights and biases.
Then, as shown in Eqn. (\ref{celltran}), (\ref{cellstate}) and (\ref{hiddenstate}), the forget gate can choose to forget the unimportant information in the previous cell state, and the input gate can retain the important information of the input data, so as to obtain the new cell state $c_t$.
Finally, the output gate determines the important information that the current cell state can output, and use the new hidden state $h_t$ as the output of LSTM.

\begin{eqnarray}
    f_t & = & \sigma(\bm{W^f}[h_{t-1},z_t] + \bm{b^f})
    \label{fgate} \\
    i_t & = & \sigma(\bm{W^i}[h_{t-1},z_t] + \bm{b^i})
    \label{igate}  \\
    o_t & = & \sigma(\bm{W^o}[h_{t-1},z_t] + \bm{b^o})
    \label{ogate} \\
    \tilde{c}_t & = & \tanh(\bm{W^c}[h_{t-1},z_t] + \bm{b^c})
    \label{celltran} \\
    c_t & = & f_t * c_{t-1} + i_t * \tilde{c}_t
    \label{cellstate} \\
    h_t & = & o_t * \tanh(c_t)
    \label{hiddenstate}
\end{eqnarray}

\paragraph{Scene Recognition}

\begin{figure}[ht]
    \centering
    \includegraphics[width=0.5\textwidth, height = 7cm]{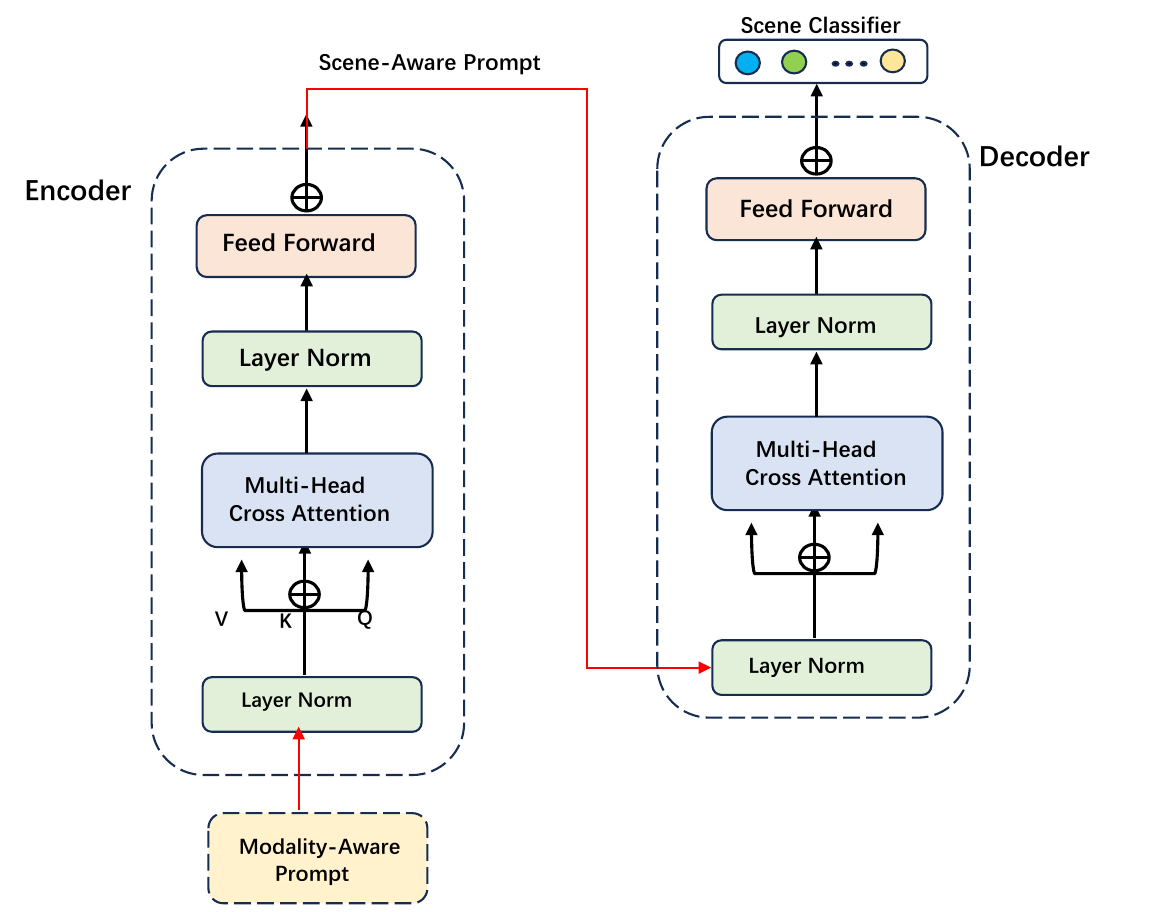}
    \caption{Network architecture of scene recognition task}
    \label{fig:scene}
\end{figure}
  
In the field of scene recognition, the accurate pairing of images and text and the effective analysis of scene-related nouns are crucial. The transformer module we designed, as shown in Fig. \ref{fig:scene}, contains encoders and decoders to meet these needs. The decoder accepts three types of input: flat image features denoted $I$, text features labeled $W$, and modality-aware prompts $\widehat{W}$. These inputs work together within the model to enhance the recognition and interpretation of the scene depicted in the image.

In the multi-head attention mechanism of our decoder, the text features $W$ act as the query $Q$. They encapsulate semantic details relevant to the scene, helping the model pinpoint image fragments that resonate with the accompanying text. These textual features are often derived from image-related descriptions, thereby directing the model's attention to the parts of the image that correspond to the textual narrative.

The modality-aware prompt $\widehat{W}$ acts as a key $K$, providing additional contextual insights. They help guide the model's focus to important image areas, especially those that are closely related to textual descriptions. Essentially, these cues act as navigation aids, directing the model's attention to the most important elements of the image.

At the same time, the image feature $I$ serves as the value $V$, which reflects the visual essence of the image. Within the multi-head attention framework, a combination of text queries and modality-aware cues are used to embed scene-related information into image features $I$. This synergy ultimately forms the scene-aware prompt, symbolized by $R$.

In the encoding stage, scene-aware prompts $R$ and text information $W$ are further refined through the attention mechanism. Here, $R$ assumes the roles of $K$ and $V$, while $W$ plays the role of $Q$. This setup enables the encoder to enhance the interaction between visual and textual elements, thereby improving the accuracy of scene understanding. The output of the encoder is an enhanced feature fusion that interweaves the visual narrative of the image with its textual description, providing a more nuanced and precise context for scene recognition. The scene recognition model's classification head then leverages this rich representation to extract indicator words that accurately describe the scene.

\subsection{Synthesis of Grid Images}

Generally, Twitter is a platform that allows everyone to show themselves from multiple aspects and angles, so we usually need to process each picture into a character-centered image. Specifically, we detect people from images using the detection algorithm GroundingDINO, and then extend a set of images to the same size through an image cropping algorithm. These steps can cover some background information in the image. On the other hand, it makes the cropped photos more beautiful.

\section{Experiment}
To demonstrate the superior performance of our model, we conducted comparative experiments to compare our method with other prompting methods and manual Twitter content generation. Ablation studies were designed to verify the importance of sub-tasks within Multi-Task Prompt Learning for generating high-quality tweets. Furthermore, a generalizability experiment was implemented to illustrate the universality of the networks employed in our sub-tasks.

\subsection{Dataset}
In the comparative experiment, both the proposed Multi-task Prompt Word Learning method and each of the comparative methods \cite{zhang2023graph, jiang2023llmlingua, li2023prompt} generated a dataset of 200 tweets. Each tweet exhibits a multimodal composition, seamlessly blending textual and visual elements. The textual component is meticulously confined to a maximum length of 40 characters. The tweet content is deliberately curated to align thematically with five distinct categories: Sports, Entertainment, Technology, Politics, and Business.

In the subsequent generalizability experiment, we utilized widely recognized public datasets. For sentiment analysis network, we employed MVSA \cite{niu2016sentiment}, Twitter15 \cite{liu2015real}, Twitter17 \cite{wang2018learning}. While Twitter15 and Twitter17 datasets comprise text and image modalities extracted from tweets, MVSA is a dataset featuring text and image modalities derived from social media posts. In the scene recognition segment, we applied Scene-15 \cite{lazebnik2006beyond}, MIT Indoor-67 \cite{quattoni2009recognizing}, and Places-205 \cite{zhou2014learning} to assess diverse scene recognition networks. Scene-15 concentrates on general outdoor and indoor scenes, MIT Indoor-67 specializes in varied indoor scenes, and Places-205 encompasses over 200 different scenes and environments, offering a comprehensive evaluation for scene recognition tasks. It's essential to emphasize that due to the absence of pertinent datasets and models exhibiting comparable performance, the generalizability experiments excluded topic classification.
\begin{table*}[!t]
   \centering
   \caption{Evaluation of Generated Social Media Content}
   \resizebox{\textwidth}{!}{
       \begin{tabular}{lcccccc}
           \toprule
           & \textbf{Manual Twitter} & \textbf{Graph-ToolFormer \cite{zhang2023graph}} & \textbf{LLM-Lingua \cite{jiang2023llmlingua}} & \textbf{Prompt Distillation \cite{li2023prompt}} & \textbf{Multi-task Prompt Words Learning (ours)} \\
           \textbf{The generated tweets} & 
            \includegraphics[width=3cm,height=3cm,keepaspectratio]{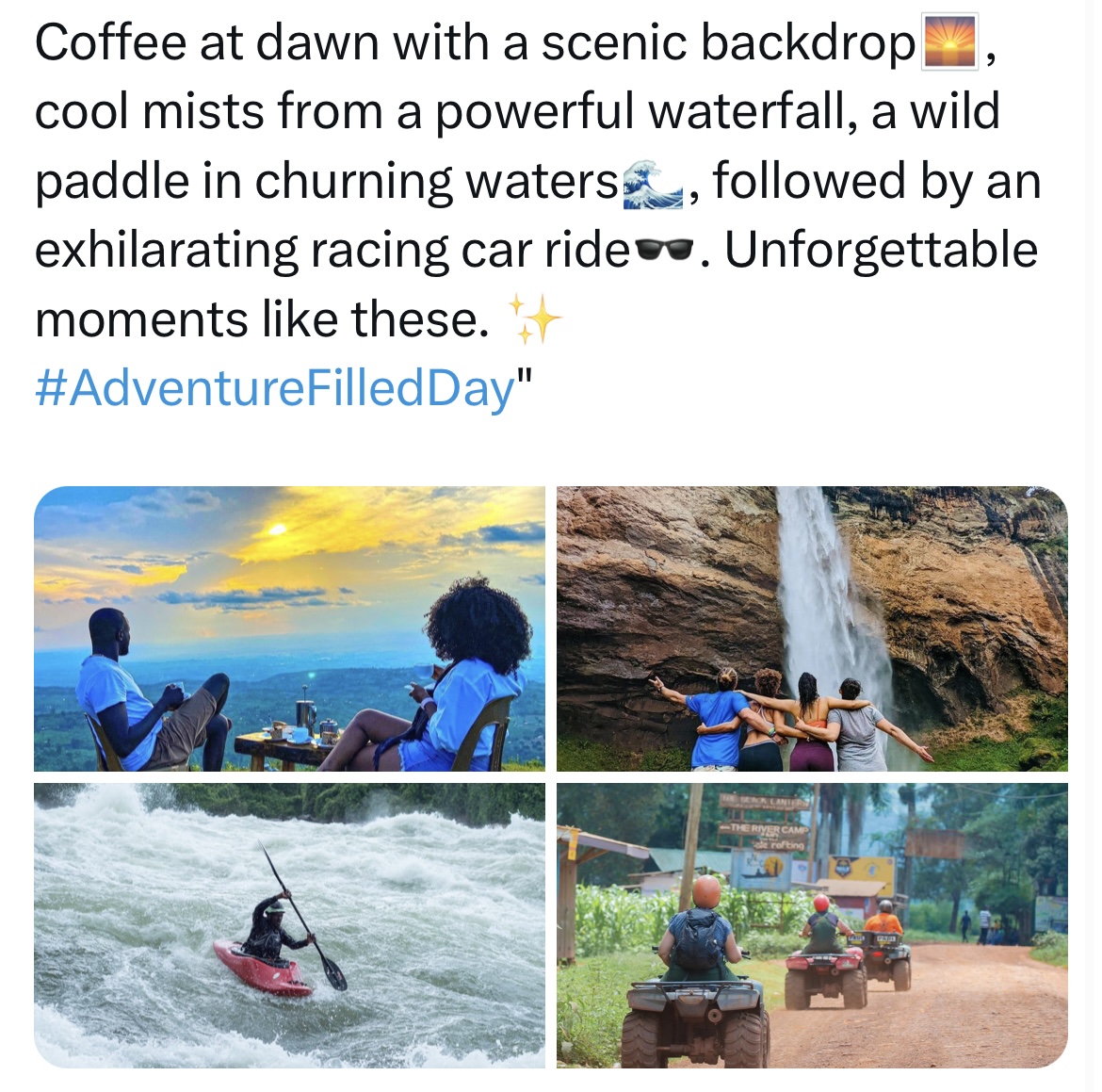} & 
            \includegraphics[width=3cm,height=3cm,keepaspectratio]{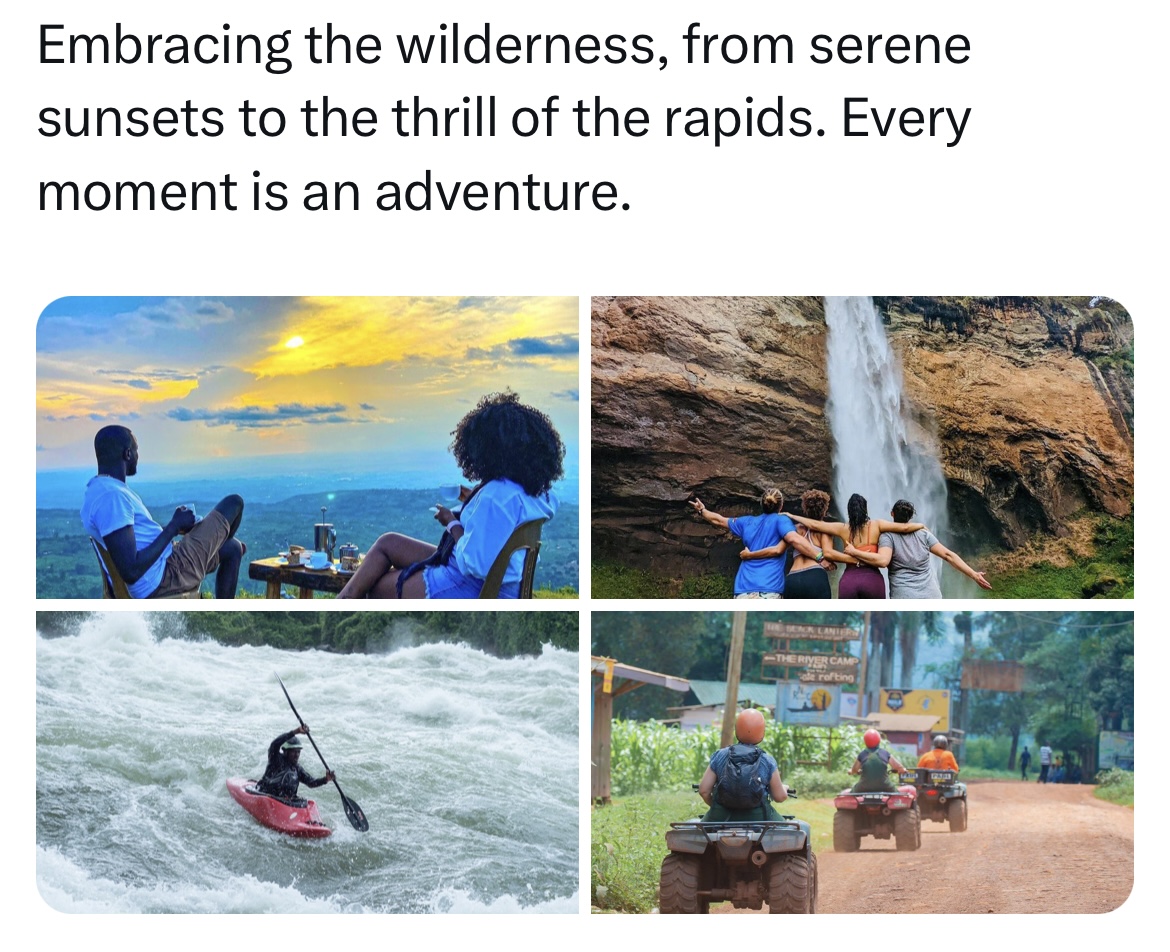} & 
            \includegraphics[width=3cm,height=3cm,keepaspectratio]{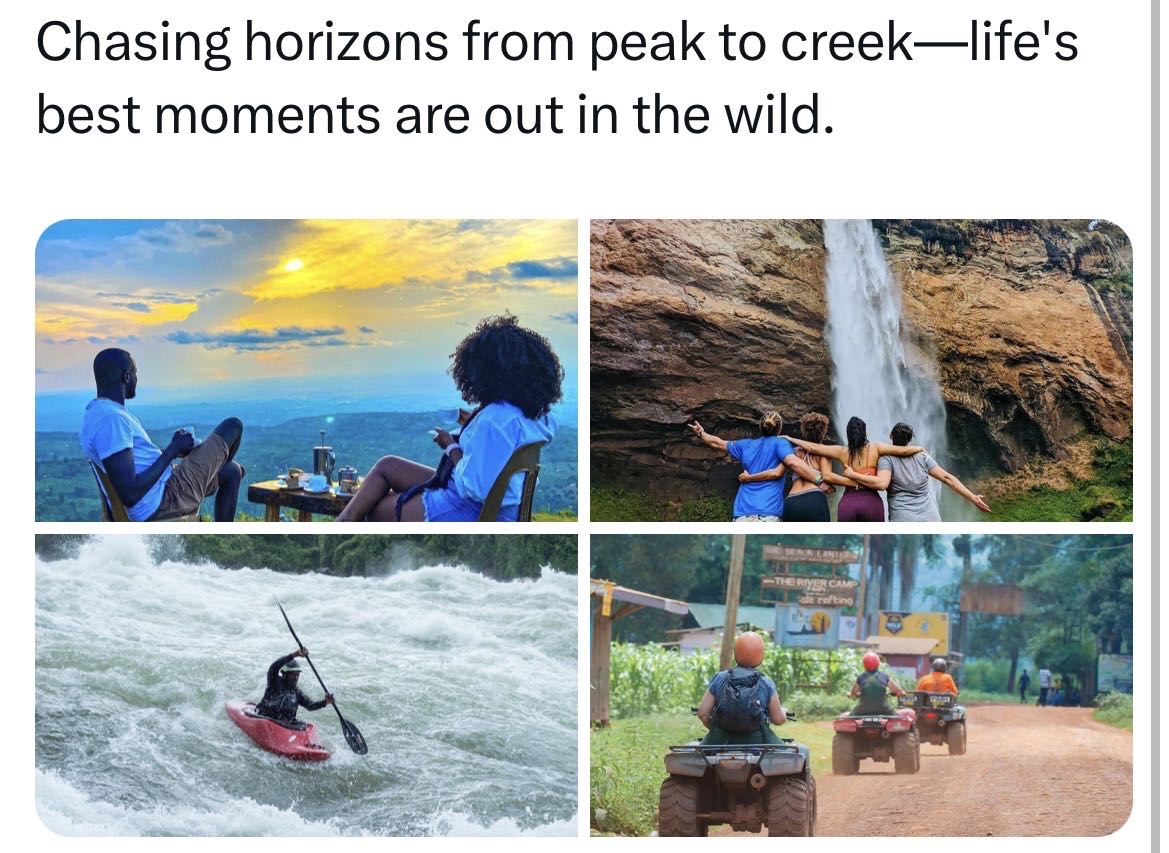} & 
            \includegraphics[width=3cm,height=3cm,keepaspectratio]{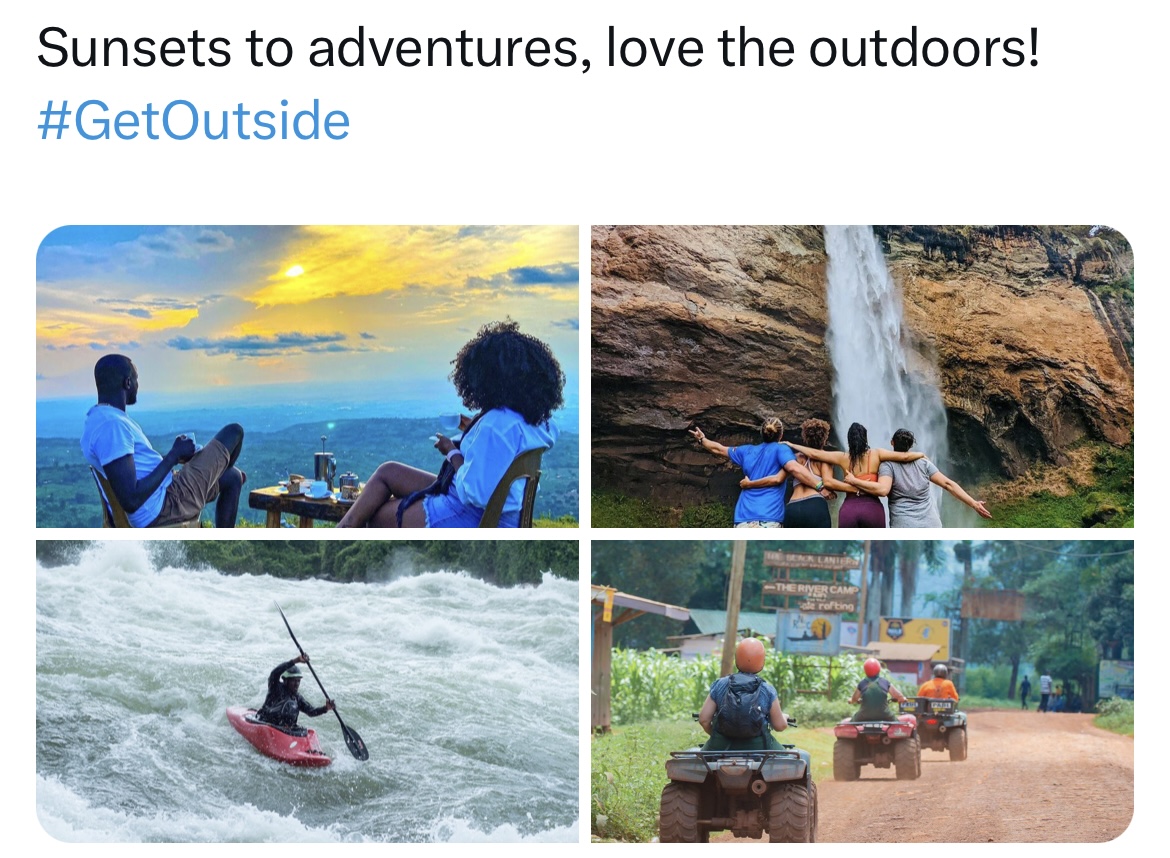} & 
            \includegraphics[width=3cm,height=3cm,keepaspectratio]{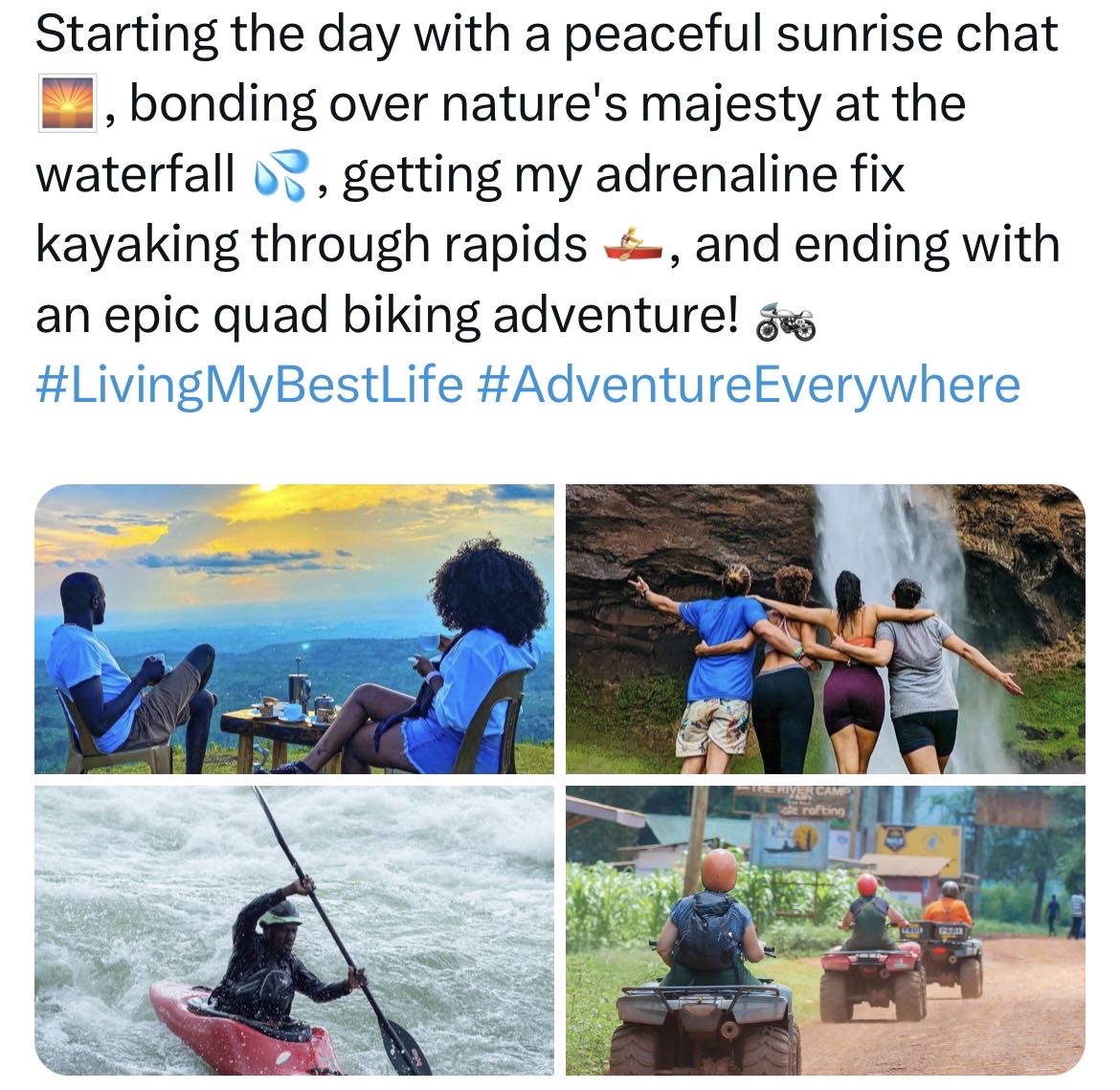} \\
            \midrule
           Relevance \& Clarity & 8.20/10 & 7.55/10 & 7.15/10 & 6.90/10 & \textbf{8.75/10} \\
           Creativity \& Originality & 7.35/10 & 5.85/10 & 5.45/10 & 4.50/10 & \textbf{7.30/10} \\
           Coherence \& Structure &7.10/10 &5.85/10 & 5.45/10& 5.30/10 & \textbf{7.60/10}\\
           Emotional Impact &6.75/10& 4.55/10 & 5.40/10& 4.40/10& \textbf{7.25/10}\\
           Engagement & 7.15/10& 5.70/10 & 5.55/10 & 5.25/10 & \textbf{7.60/10}\\
           
           Overall &7.50/10 & 5.85& 5.45& 5.35& \textbf{8.05/10}\\
           \bottomrule
       \end{tabular}
   }
\label{compare_1}
\end{table*}

\begin{table*}[!t]
\centering
\caption{Ablation Study of Multi-task Prompt Word Learning Method}
\resizebox{\textwidth}{!}{
\begin{tabular}{lcccccc}
\toprule
\textbf{Method} & \textbf{Relevance \& Clarity} & \textbf{Creativity 
 \& Originality} & \textbf{Coherence \& Structure} & \textbf{Emotional Impact} & \textbf{Engagement}  & \textbf{Overall} \\
\midrule
MPWL w/o Scene Recognition & 8.35 & 7.20 & 7.55 & 7.25 & 7.15  & 7.70 \\
MPWL w/o Topic Classification & 7.90 & 7.15 & 7.45 & 7.20 & 6.80 & 7.35 \\
MPWL w/o Sentiment Analyse  & 7.75 & 7.15 & 7.40 & 6.15 & 6.60 &  7.15 \\ 
w/o MPWL & 6.60 & 4.35 & 5.20 & 5.05 & 5.15 &  5.20 \\
\hline
Ours (with MPWL) & \textbf{8.75} & \textbf{7.30} & \textbf{7.60} & \textbf{7.25} & \textbf{7.60}  & \textbf{8.05} \\
\bottomrule
\end{tabular}}
\label{compare2}
\end{table*}
\subsection{Implementation Details}
We utilize the unified GPT-4 version \cite{openai2023gpt4} by OpenAI without further branching. For Multi-task Prompt Word Learning, the topic classification model undergoes 500 epochs with a batch size of 4 and a learning rate of 0.005. Simultaneously, the sentiment analysis network is refined over 500 epochs on the Twitter-15 dataset with a batch size of 16 and a consistent learning rate of 0.005. The dedicated scene recognition network is trained on the Scene-15 dataset for 300 epochs, utilizing a batch size of 16 and a slightly increased learning rate of 0.008. All models leverage a single NVIDIA 3090 GPU throughout training.

\begin{figure}[ht]
    \centering
    \includegraphics[width=0.48\textwidth, height = 9cm]{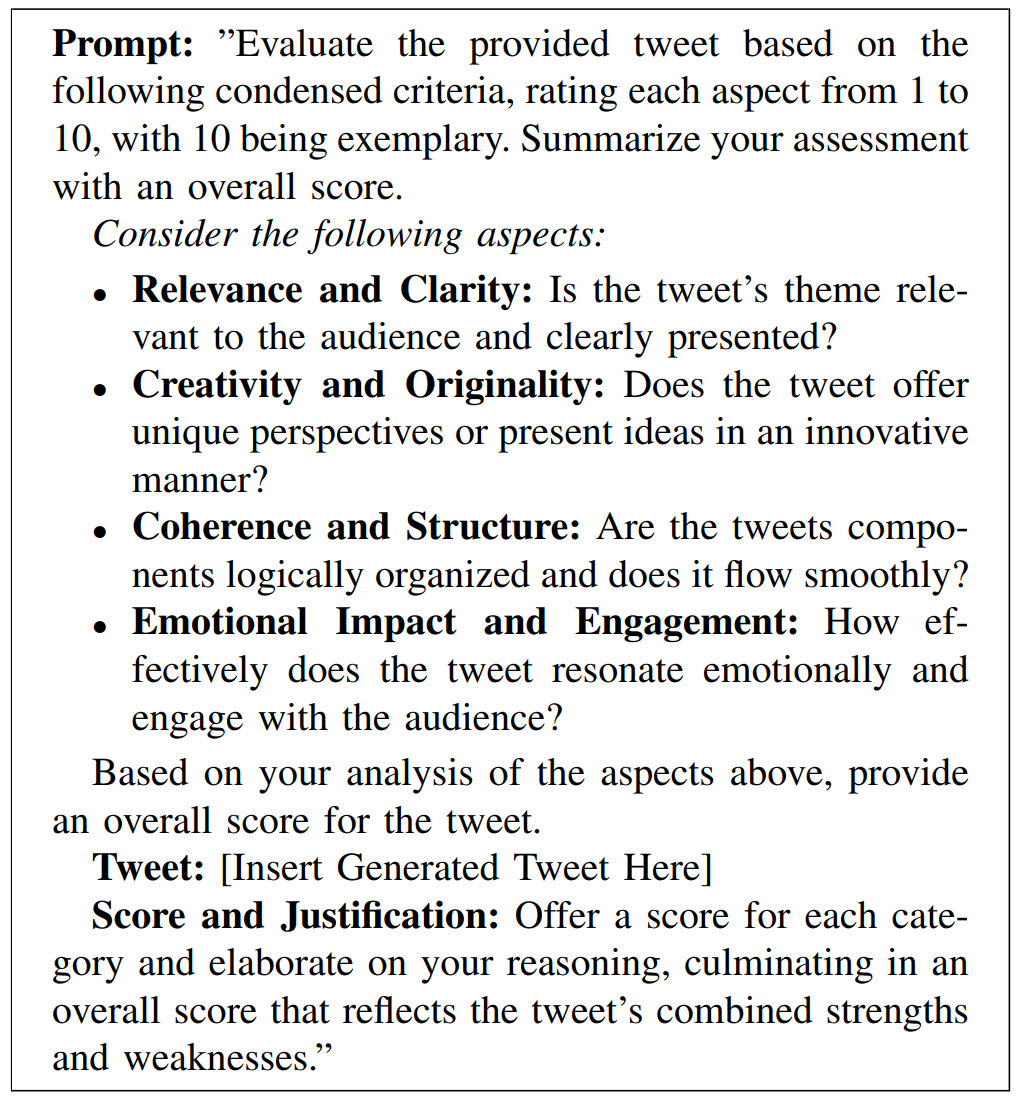}
    \caption{The Designed Template for Scoring Generated Tweets}
    \label{prompting2}
\end{figure}

In the comparative experiment, we have designed a template to facilitate GPT in scoring tweets. As illustrated in Figure \ref{prompting2}, the evaluation of generated tweet content encompasses four main aspects: Relevance and Clarity, Creativity and Originality, Coherence and Structure, and Emotional Impact and Engagement. Each aspect is rated on a scale from 1 to 10. This approach aims to reduce biases and variations typically found in human scoring methods. In the ablation study, this template is similarly employed, but with the addition of blanks to account for absent subtasks.

For the generalizability experiment, we isolate sub-networks in Multi-Task Prompt Learning and compare their performance with the corresponding task's network on universal datasets. We use Accuracy and F1-score to measure the performance of different methods. 

\subsection{Comparative Experiments}

As shown in Table \ref{compare_1}, the quality of tweets generated under the guidance of Multi-task Prompt Word Learning (MPWL) is relatively higher compared to manually crafted tweets. MPWL outperforms in almost all evaluated aspects, except for Creativity and Originality, where it slightly falls manually crafted tweets. The lesser performance of other prompting learning methods \cite{zhang2023graph, jiang2023llmlingua, li2023prompt} can be linked to their unsuitability for social media content creation. These methods \cite{zhang2023graph, jiang2023llmlingua, li2023prompt} lack the specialized features necessary to extract themes, emotions, and locations from images, and they do not have structured templates crucial for crafting standard tweets. As a result, their tweets often lack the emotional depth, creativity, and overall quality that are typical of manually crafted tweets. In contrast, MPWL shows more skill in these areas.

\subsection{Ablation study}
To further validate the effectiveness of proposed Multi-task Prompt Word Learning (MPWL), we conducted ablation experiments with different settings. Specifically, we performed experiment by removing the MPWL, denoted as ``w/o MPWL" in Table \ref{compare2}. Additionally, experiments were conducted by removing each component from the MPWL, labeled as ``MPWL w/o Scene Recognition", ``MPWL w/o Topic Classification", or ``MPWL w/o Sentiment Analyse". The results reveal that: (1) MPWL with all components achieves the best performance on generated content, while removing any module results in sub-optimal results; (2) ``w/o MPWL" performs worse than MPWL, highlighting the effectiveness of  scene recognition, topic classification and sentiment analysis in enhancing the overall quality of the content.
\subsection{Generalizability Experiments}
To further validate the generalizability of Multi-Task Media Prompting method, we test its sub-network in different tasks, including sentiment analysis, and scene recognition. This diverse range of tasks provided a comprehensive evaluation of its adaptability. As shown in Table \ref{tab:sentiment} \ref{scene_recogntion},
the results confirmed that Multi-Task Media Prompting not only performs on par with the most commonly used existing methods \cite{cai2015convolutional, yuadapting,yu2019adapting, yang2015multi, xie2015hybrid, herranz2016scene} but also showcases its superiority in managing diverse data types and meeting different analytical requirements. This extension into various tasks emphasized the flexibility and effectiveness of Multi-Task Media Prompting in a wide range of applications.
\begin{table}[H]
\centering
\scriptsize
\caption{Performance comparison (Accuracy(\%) and F1(\%)) of different Sentiment analysis models.}
\begin{tabular}{lcccccc}
\hline
 & \multicolumn{2}{c}{\textbf{MVSA}} & \multicolumn{2}{c}{\textbf{Twitter15}} & \multicolumn{2}{c}{\textbf{Twitter17}} \\
\hline
\textbf{Model} & ACC $\uparrow$ & F1 $\uparrow$ & ACC $\uparrow$ & F1 $\uparrow$ & ACC $\uparrow$ & F1 $\uparrow$ \\
\hline
CNN-Multi \cite{cai2015convolutional} & 66.30 & 64.19 & 64.38 & 63.48 & 61.87 & 60.26 \\
mBERT(BOTH) \cite{yuadapting} & 68.46 & 66.83 & 75.31 & 70.18 & 69.61 & 67.12 \\
TomBert \cite{yu2019adapting} & \textbf{69.86} & 65.84 & 76.18 & 71.27 & \textbf{70.50} & 68.04 \\
Ours & 69.12 & \textbf{66.56} & \textbf{76.31} & \textbf{71.56} & 70.21 & \textbf{68.43} \\
\hline
\end{tabular}
\label{tab:sentiment}
\end{table}

\begin{table}[H]
\centering
\scriptsize
\caption{Performance comparison (Accuracy(\%) and F1(\%)) of different Scene recognition models.}
\begin{tabular}{lcccccc}
\hline
 & \multicolumn{2}{c}{\textbf{Scene-15}} & \multicolumn{2}{c}{\textbf{MIT Indoor-67}} & \multicolumn{2}{c}{\textbf{Places-205}} \\
\hline
\textbf{Model} & ACC $\uparrow$ & F1 $\uparrow$ & ACC $\uparrow$ & F1 $\uparrow$ & ACC $\uparrow$ & F1 $\uparrow$ \\
\hline
DAG-CNN \cite{yang2015multi} & 92.90 & 89.41 & 77.50 & 73.41 & 56.20 & 52.28 \\
Hybrid CNNs \cite{xie2015hybrid} & 91.21 & 88.86 & 82.24 & 79.56 & 64.53 & 61.89 \\
Multi-scale CNNs \cite{herranz2016scene}& 95.18 & 92.87 & 86.04 & 82.69 & 70.17 & \textbf{67.56} \\
Ours & \textbf{95.34} & \textbf{93.17} & \textbf{86.07} & \textbf{82.86} & \textbf{70.23} & 67.71 \\
\hline
\end{tabular}
\label{scene_recogntion}
\end{table}

\section{Conclusion}

In our work, we apply artificial intelligence technology to content generation for social media and propose a prompt word generation framework based on multi-modal information fusion, which uses topic classification, sentiment analysis, and scene recognition. Generate more comprehensive prompt words to generate high-quality tweets through ChatGPT. Furthermore, in the absence of effective and objective evaluation methods in the field of content generation, we propose to use the ChatGPT tool to achieve large-scale evaluation of content generation methods. The final results show that our approach produces higher-quality tweets compared to other on-the-fly learning methods and manually crafted posts. In future work, we will focus on further processing and beautification of images.

\newpage

\bibliographystyle{IEEEtran}
\bibliography{IEEEtran}

\begin{thebibliography}{10}
\providecommand{\url}[1]{#1}
\csname url@samestyle\endcsname
\providecommand{\newblock}{\relax}
\providecommand{\bibinfo}[2]{#2}
\providecommand{\BIBentrySTDinterwordspacing}{\spaceskip=0pt\relax}
\providecommand{\BIBentryALTinterwordstretchfactor}{4}
\providecommand{\BIBentryALTinterwordspacing}{\spaceskip=\fontdimen2\font plus
\BIBentryALTinterwordstretchfactor\fontdimen3\font minus \fontdimen4\font\relax}
\providecommand{\BIBforeignlanguage}[2]{{%
\expandafter\ifx\csname l@#1\endcsname\relax
\typeout{** WARNING: IEEEtran.bst: No hyphenation pattern has been}%
\typeout{** loaded for the language `#1'. Using the pattern for}%
\typeout{** the default language instead.}%
\else
\language=\csname l@#1\endcsname
\fi
#2}}
\providecommand{\BIBdecl}{\relax}
\BIBdecl

\bibitem{kaplan2010users}
A.~M. Kaplan and M.~Haenlein, ``Users of the world, unite! the challenges and opportunities of social media,'' \emph{Business horizons}, vol.~53, no.~1, pp. 59--68, 2010.

\bibitem{alves2016social}
H.~Alves, C.~Fernandes, and M.~Raposo, ``Social media marketing: a literature review and implications,'' \emph{Psychology \& Marketing}, vol.~33, no.~12, pp. 1029--1038, 2016.

\bibitem{davis2022investigating}
G.~Davis, M.~Grierson \emph{et~al.}, ``Investigating attitudes of professional writers to gpt text generation ai based creative support tools,'' \emph{.}, 2022.

\bibitem{carr2015social}
C.~T. Carr and R.~A. Hayes, ``Social media: Defining, developing, and divining,'' \emph{Atlantic journal of communication}, vol.~23, no.~1, pp. 46--65, 2015.

\bibitem{liu2023grounding}
S.~Liu, Z.~Zeng, T.~Ren, F.~Li, H.~Zhang, J.~Yang, C.~Li, J.~Yang, H.~Su, J.~Zhu \emph{et~al.}, ``Grounding dino: Marrying dino with grounded pre-training for open-set object detection,'' \emph{arXiv preprint arXiv:2303.05499}, 2023.

\bibitem{yunjiu2022artificial}
L.~Yunjiu, W.~Wei, and Y.~Zheng, ``Artificial intelligence-generated and human expert-designed vocabulary tests: A comparative study,'' \emph{SAGE Open}, vol.~12, no.~1, p. 21582440221082130, 2022.

\bibitem{karras2017progressive}
T.~Karras, T.~Aila, S.~Laine, and J.~Lehtinen, ``Progressive growing of gans for improved quality, stability, and variation,'' \emph{arXiv preprint arXiv:1710.10196}, 2017.

\bibitem{huang2018multimodal}
X.~Huang, M.-Y. Liu, S.~Belongie, and J.~Kautz, ``Multimodal unsupervised image-to-image translation,'' in \emph{Proceedings of the European conference on computer vision (ECCV)}, 2018, pp. 172--189.

\bibitem{brown2020language}
T.~Brown, B.~Mann, N.~Ryder, M.~Subbiah, J.~D. Kaplan, P.~Dhariwal, A.~Neelakantan, P.~Shyam, G.~Sastry, A.~Askell \emph{et~al.}, ``Language models are few-shot learners,'' \emph{Advances in neural information processing systems}, vol.~33, pp. 1877--1901, 2020.

\bibitem{chowdhery2023palm}
A.~Chowdhery, S.~Narang, J.~Devlin, M.~Bosma, G.~Mishra, A.~Roberts, P.~Barham, H.~W. Chung, C.~Sutton, S.~Gehrmann \emph{et~al.}, ``Palm: Scaling language modeling with pathways,'' \emph{Journal of Machine Learning Research}, vol.~24, no. 240, pp. 1--113, 2023.

\bibitem{chung2022scaling}
H.~W. Chung, L.~Hou, S.~Longpre, B.~Zoph, Y.~Tay, W.~Fedus, E.~Li, X.~Wang, M.~Dehghani, S.~Brahma \emph{et~al.}, ``Scaling instruction-finetuned language models,'' \emph{arXiv preprint arXiv:2210.11416}, 2022.

\bibitem{poesia2022synchromesh}
G.~Poesia, O.~Polozov, V.~Le, A.~Tiwari, G.~Soares, C.~Meek, and S.~Gulwani, ``Synchromesh: Reliable code generation from pre-trained language models,'' \emph{arXiv preprint arXiv:2201.11227}, 2022.

\bibitem{brooks2023instructpix2pix}
T.~Brooks, A.~Holynski, and A.~A. Efros, ``Instructpix2pix: Learning to follow image editing instructions,'' in \emph{Proceedings of the IEEE/CVF Conference on Computer Vision and Pattern Recognition}, 2023, pp. 18\,392--18\,402.

\bibitem{schick2023toolformer}
T.~Schick, J.~Dwivedi-Yu, R.~Dess{\`\i}, R.~Raileanu, M.~Lomeli, L.~Zettlemoyer, N.~Cancedda, and T.~Scialom, ``Toolformer: Language models can teach themselves to use tools,'' \emph{arXiv preprint arXiv:2302.04761}, 2023.

\bibitem{peng2023instruction}
B.~Peng, C.~Li, P.~He, M.~Galley, and J.~Gao, ``Instruction tuning with gpt-4,'' \emph{arXiv preprint arXiv:2304.03277}, 2023.

\bibitem{openai2023gpt4}
OpenAI, ``Gpt-4 technical report,'' 2023.

\bibitem{gan2022vision}
Z.~Gan, L.~Li, C.~Li, L.~Wang, Z.~Liu, J.~Gao \emph{et~al.}, ``Vision-language pre-training: Basics, recent advances, and future trends,'' \emph{Foundations and Trends{\textregistered} in Computer Graphics and Vision}, vol.~14, no. 3--4, pp. 163--352, 2022.

\bibitem{liu2023pre}
P.~Liu, W.~Yuan, J.~Fu, Z.~Jiang, H.~Hayashi, and G.~Neubig, ``Pre-train, prompt, and predict: A systematic survey of prompting methods in natural language processing,'' \emph{ACM Computing Surveys}, vol.~55, no.~9, pp. 1--35, 2023.

\bibitem{radford2021learning}
A.~Radford, J.~W. Kim, C.~Hallacy, A.~Ramesh, G.~Goh, S.~Agarwal, G.~Sastry, A.~Askell, P.~Mishkin, J.~Clark \emph{et~al.}, ``Learning transferable visual models from natural language supervision,'' in \emph{International Conference on Machine Learning}.\hskip 1em plus 0.5em minus 0.4em\relax PMLR, 2021, pp. 8748--8763.

\bibitem{zhou2022learning}
K.~Zhou, J.~Yang, C.~C. Loy, and Z.~Liu, ``Learning to prompt for vision-language models,'' \emph{International Journal of Computer Vision}, vol. 130, no.~9, pp. 2337--2348, 2022.

\bibitem{zhou2022conditional}
------, ``Conditional prompt learning for vision-language models,'' in \emph{Proceedings of the IEEE/CVF Conference on Computer Vision and Pattern Recognition}, 2022, pp. 16\,816--16\,825.

\bibitem{lu2022prompt}
Y.~Lu, J.~Liu, Y.~Zhang, Y.~Liu, and X.~Tian, ``Prompt distribution learning,'' in \emph{Proceedings of the IEEE Conference on Computer Vision and Pattern Recognition}, 2022, pp. 5206--5215.

\bibitem{zhu2023prompt}
B.~Zhu, Y.~Niu, Y.~Han, Y.~Wu, and H.~Zhang, ``Prompt-aligned gradient for prompt tuning,'' in \emph{Proceedings of the IEEE International Conference on Computer Vision}, 2023, pp. 15\,659--15\,669.

\bibitem{zhang2023graph}
J.~Zhang, ``Graph-toolformer: To empower llms with graph reasoning ability via prompt augmented by chatgpt,'' \emph{arXiv preprint arXiv:2304.11116}, 2023.

\bibitem{jiang2023llmlingua}
H.~Jiang, Q.~Wu, C.-Y. Lin, Y.~Yang, and L.~Qiu, ``Llmlingua: Compressing prompts for accelerated inference of large language models,'' \emph{arXiv preprint arXiv:2310.05736}, 2023.

\bibitem{li2023prompt}
L.~Li, Y.~Zhang, and L.~Chen, ``Prompt distillation for efficient llm-based recommendation,'' in \emph{Proceedings of the 32nd ACM International Conference on Information and Knowledge Management}, 2023, pp. 1348--1357.

\bibitem{rao2022denseclip}
Y.~Rao, W.~Zhao, G.~Chen, Y.~Tang, Z.~Zhu, G.~Huang, J.~Zhou, and J.~Lu, ``Denseclip: Language-guided dense prediction with context-aware prompting,'' in \emph{Proceedings of the IEEE Conference on Computer Vision and Pattern Recognition}, 2022, pp. 18\,082--18\,091.

\bibitem{gao2023clip}
P.~Gao, S.~Geng, R.~Zhang, T.~Ma, R.~Fang, Y.~Zhang, H.~Li, and Y.~Qiao, ``Clip-adapter: Better vision-language models with feature adapters,'' \emph{International Journal of Computer Vision}, pp. 1--15, 2023.

\bibitem{shen2023hugginggpt}
Y.~Shen, K.~Song, X.~Tan, D.~Li, W.~Lu, and Y.~Zhuang, ``Hugginggpt: Solving ai tasks with chatgpt and its friends in huggingface,'' \emph{arXiv preprint arXiv:2303.17580}, 2023.

\bibitem{khan2022transformers}
S.~Khan, M.~Naseer, M.~Hayat, S.~W. Zamir, F.~S. Khan, and M.~Shah, ``Transformers in vision: A survey,'' \emph{ACM computing surveys (CSUR)}, vol.~54, no. 10s, pp. 1--41, 2022.

\bibitem{lin2014microsoft}
T.-Y. Lin, M.~Maire, S.~Belongie, J.~Hays, P.~Perona, D.~Ramanan, P.~Doll{\'a}r, and C.~L. Zitnick, ``Microsoft coco: Common objects in context,'' in \emph{Computer Vision--ECCV 2014: 13th European Conference, Zurich, Switzerland, September 6-12, 2014, Proceedings, Part V 13}.\hskip 1em plus 0.5em minus 0.4em\relax Springer, 2014, pp. 740--755.

\bibitem{Li_2022_CVPR}
M.~Li, R.~Xu, S.~Wang, L.~Zhou, X.~Lin, C.~Zhu, M.~Zeng, H.~Ji, and S.-F. Chang, ``Clip-event: Connecting text and images with event structures,'' in \emph{Proceedings of the IEEE/CVF Conference on Computer Vision and Pattern Recognition (CVPR)}, June 2022, pp. 16\,420--16\,429.

\bibitem{manning2014stanford}
C.~D. Manning, M.~Surdeanu, J.~Bauer, J.~R. Finkel, S.~Bethard, and D.~McClosky, ``The stanford corenlp natural language processing toolkit,'' in \emph{Proceedings of 52nd annual meeting of the association for computational linguistics: system demonstrations}, 2014, pp. 55--60.

\bibitem{devlin2018bert}
J.~Devlin, M.-W. Chang, K.~Lee, and K.~Toutanova, ``Bert: Pre-training of deep bidirectional transformers for language understanding,'' \emph{arXiv preprint arXiv:1810.04805}, 2018.

\bibitem{han2022survey}
K.~Han, Y.~Wang, H.~Chen, X.~Chen, J.~Guo, Z.~Liu, Y.~Tang, A.~Xiao, C.~Xu, Y.~Xu \emph{et~al.}, ``A survey on vision transformer,'' \emph{IEEE transactions on pattern analysis and machine intelligence}, vol.~45, no.~1, pp. 87--110, 2022.

\bibitem{niu2016sentiment}
T.~Niu, S.~Zhu, L.~Pang, and A.~El~Saddik, ``Sentiment analysis on multi-view social data,'' in \emph{MultiMedia Modeling: 22nd International Conference, MMM 2016, Miami, FL, USA, January 4-6, 2016, Proceedings, Part II 22}.\hskip 1em plus 0.5em minus 0.4em\relax Springer, 2016, pp. 15--27.

\bibitem{liu2015real}
X.~Liu, A.~Nourbakhsh, Q.~Li, R.~Fang, and S.~Shah, ``Real-time rumor debunking on twitter,'' in \emph{Proceedings of the 24th ACM international on conference on information and knowledge management}, 2015, pp. 1867--1870.

\bibitem{wang2018learning}
B.~Wang and W.~Lu, ``Learning latent opinions for aspect-level sentiment classification,'' in \emph{Proceedings of the AAAI Conference on Artificial Intelligence}, vol.~32, no.~1, 2018.

\bibitem{lazebnik2006beyond}
S.~Lazebnik, C.~Schmid, and J.~Ponce, ``Beyond bags of features: Spatial pyramid matching for recognizing natural scene categories,'' in \emph{2006 IEEE computer society conference on computer vision and pattern recognition (CVPR'06)}, vol.~2.\hskip 1em plus 0.5em minus 0.4em\relax IEEE, 2006, pp. 2169--2178.

\bibitem{quattoni2009recognizing}
A.~Quattoni and A.~Torralba, ``Recognizing indoor scenes,'' in \emph{2009 IEEE conference on computer vision and pattern recognition}.\hskip 1em plus 0.5em minus 0.4em\relax IEEE, 2009, pp. 413--420.

\bibitem{zhou2014learning}
B.~Zhou, A.~Lapedriza, J.~Xiao, A.~Torralba, and A.~Oliva, ``Learning deep features for scene recognition using places database,'' \emph{Advances in neural information processing systems}, vol.~27, 2014.

\bibitem{cai2015convolutional}
G.~Cai and B.~Xia, ``Convolutional neural networks for multimedia sentiment analysis,'' in \emph{Natural Language Processing and Chinese Computing: 4th CCF Conference, NLPCC 2015, Nanchang, China, October 9-13, 2015, Proceedings 4}.\hskip 1em plus 0.5em minus 0.4em\relax Springer, 2015, pp. 159--167.

\bibitem{yuadapting}
J.~Yu and J.~Jiang, ``Adapting bert for target-oriented multimodal sentiment classification,'' in \emph{Proceedings of the Twenty-Eighth International Joint Conference on Artificial Intelligence}.\hskip 1em plus 0.5em minus 0.4em\relax IJCAI, 2019.

\bibitem{yu2019adapting}
------, ``Adapting bert for target-oriented multimodal sentiment classification,'' in \emph{Proceedings of the Twenty-Eighth International Joint Conference on Artificial Intelligence}.\hskip 1em plus 0.5em minus 0.4em\relax IJCAI, 2019.

\bibitem{yang2015multi}
S.~Yang and D.~Ramanan, ``Multi-scale recognition with dag-cnns,'' in \emph{Proceedings of the IEEE international conference on computer vision}, 2015, pp. 1215--1223.

\bibitem{xie2015hybrid}
G.-S. Xie, X.-Y. Zhang, S.~Yan, and C.-L. Liu, ``Hybrid cnn and dictionary-based models for scene recognition and domain adaptation,'' \emph{IEEE Transactions on Circuits and Systems for Video Technology}, vol.~27, no.~6, pp. 1263--1274, 2015.

\bibitem{herranz2016scene}
L.~Herranz, S.~Jiang, and X.~Li, ``Scene recognition with cnns: objects, scales and dataset bias,'' in \emph{Proceedings of the IEEE Conference on Computer Vision and Pattern Recognition}, 2016, pp. 571--579.

\end{thebibliography}

\end{document}